\documentclass[runningheads]{llncs}

\usepackage{eccv}

\usepackage{eccvabbrv}

\usepackage{wrapfig}
\usepackage{graphicx}
\usepackage{booktabs}
\usepackage{multirow}
\usepackage{pifont}
\usepackage[table]{xcolor}
\usepackage{algorithm}
\usepackage{algorithmic}
\usepackage{booktabs}

\usepackage[accsupp]{axessibility}  

\usepackage{hyperref}

\usepackage{orcidlink}

\begin{document}

\title{Robustness Emerges Early in Training Dynamics, but Is Not Preserved} 

\titlerunning{Robustness Emerges Early in Training Dynamics, but Is Not Preserved}

\author{Jiangang Yang\inst{1}\orcidlink{0000-0002-0464-8336} \and
Wenhui Shi\inst{1,2}\orcidlink{0009-0009-1662-7151} \and
Lu Hu\inst{1,2}\orcidlink{0009-0009-0792-6360} \and
Jing Xing\inst{1}\orcidlink{0009-0009-9607-3987} \and
Jian Liu\inst{1}}

\authorrunning{J. Yang et al.}

\institute{Institute of Microelectronics, Chinese Academy of Sciences, Beijing, China \and
University of Chinese Academy of Sciences, Beijing, China\\
\email{\{yangjiangang, liujian\}@ime.ac.cn}}

\maketitle

\begin{abstract}
Robustness to natural corruptions remains a fundamental challenge for deep neural networks. In this paper, we identify a robustness fading phenomenon where shallow layers spontaneously develop robust representations and flat loss landscapes in early training, yet these properties are not preserved during standard convergence. To address this, we propose a framework that performs strategic interventions on training dynamics to stabilize the empirically identified early-emergent robust priors. Our approach includes two parameter-free strategies: Early-Phase Stabilization~(EPS) and Asymmetric Weight Reversion~(AWR), which stabilize or recover robust shallow configurations without modifying the model architecture or introducing learnable parameters. Extensive experiments demonstrate the efficacy of our framework across various benchmarks and architectures, yielding significant gains in downstream transfer, dynamic adaptation, and diverse computer vision applications.
\end{abstract}

\section{Introduction}
Deep neural networks have achieved remarkable success across a wide range of vision tasks, yet their robustness remains a critical concern~\cite{zhao2022ood,mayilvahanan2024does,liu2025comprehensive}. Prior work on robustness mainly follows two directions: robustness to adversarial perturbations~\cite{yuan2019adversarial}, which are small, intentionally designed input modifications that induce mispredictions, and robustness to natural corruptions such as noise, blur, and contrast changes~\cite{hendrycks2019benchmarking}. Unlike adversarial examples, natural corruptions arise from realistic variations in data acquisition and environmental conditions. Autonomous systems are affected by fog or low illumination, while medical imaging systems experience systematic degradations due to differences in scanners, acquisition protocols, or patient motion. Improving robustness to such naturally occurring corruptions is essential for reliable deployment in safety-critical applications.

\begin{figure}[tb] 
  \centering
  \includegraphics[width=0.95\linewidth]{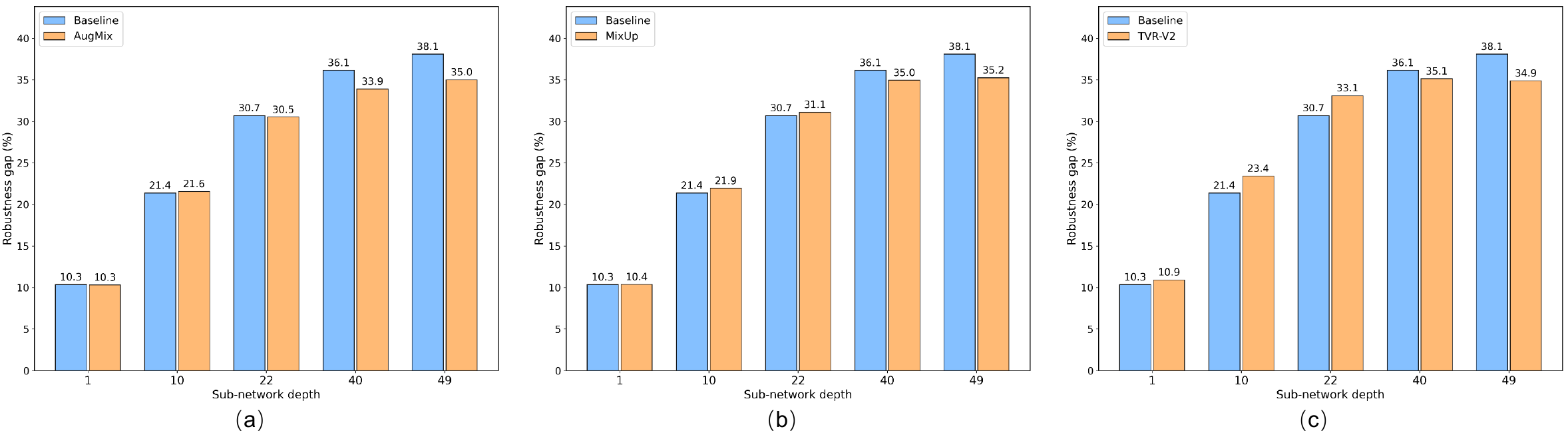} 
  \caption{Robustness gap analysis across subnetworks. We compare ResNet-50 with (a) AugMix~\cite{hendrycksaugmix}, (b) Mixup~\cite{zhang2018mixup}, and (c) the TVR-V2 training recipe~\cite{vryniotis2021train}. While all three methods significantly reduce the robustness gap in deeper subnetworks~(e.g., \#40 and \#49), their performance in shallow subnetworks~(e.g., \#1 and \#10) remains nearly identical to the baseline, failing to deliver meaningful gains.}
  \label{fig:robustness_gap_analysis}
\end{figure}

To enhance robustness to natural corruptions, prior work has explored diverse strategies, including data augmentation~\cite{hendrycksaugmix,qin2022understanding,modas2022prime,vaish2024fourier}, regularization~\cite{foret2020sharpness,trinh2024improving}, composite training recipes~\cite{wightman2021resnet,vryniotis2021train}, ensembling~\cite{saikia2021improving,diffenderfer2021winning}, and bio-inspired architectures~\cite{dapello2020simulating,vryniotis2021train}. However, as shown in \cref{fig:robustness_gap_analysis}, our layer-wise linear probing analysis~\cite{alain2016understanding}, which measures the performance gap between clean and corrupted inputs, reveals that even state-of-the-art methods deliver gains almost entirely in deeper layers, with shallow layers remaining essentially unchanged. This imbalance suggests that shallow-layer robustness is a key bottleneck. Bio-inspired work often addresses this by augmenting the network front-end with explicit architectural priors; for example, VOneNet~\cite{dapello2020simulating} inserts a V1-inspired shallow module and achieves substantial robustness gains, implying that standard training under-utilizes robustness capacity in early layers. 
This leads to a critical consideration: if robustness can be hard-coded into early layers, it may as well emerge spontaneously through the learning process. Current empirical evidence supports the possibility that training dynamics play a crucial role. For instance, studies on robust pruning indicate~\cite{diffenderfer2021winning} that high-performance subnetworks exist within over-parameterized models, while the phenomenon of robust overfitting~\cite{rice2020overfitting} indicates that adversarial robustness can peak early and then decline as training progresses. This paradigm shift from static architecture to temporal dynamics motivates a fundamental inquiry: \textit{under natural corruptions, do training dynamics give rise to a robust shallow subnetwork early on, but standard training fails to preserve it?}

Motivated by this inquiry, we dissect the training dynamics of shallow subnetworks. By tracing weight trajectories, we find that shallow layers undergo an active exploration phase in early training rather than rapid convergence. This raises the hypothesis that a more robust shallow subnetwork emerges during this exploratory stage but fails to persist through standard training. To examine this, our analysis of representation stability uncovers a striking degradation: shallow features are most consistent and information-preserving under corruptions early on, yet this implicit robustness fades toward convergence. We further connect this early-to-late shift to the model's optimization geometry under input perturbations. Specifically, by freezing shallow subnetworks from various training epochs and retraining the remaining layers, we observe that early-phase structures sustain a significantly flatter loss surface under perturbations, a key indicator of enhanced robustness. Conversely, subnetworks from later epochs induce landscape sharpening. Collectively, these results suggest that standard training dynamics do not retain early robustness-relevant properties, motivating training-time mechanisms to preserve them.

Building on these findings, we propose a framework that intervenes in training dynamics to preserve early-emergent robustness. This framework provides two simple, easy-to-implement strategies. Early-Phase Stabilization~(EPS) halts the shallow subnetwork updates early in training to lock in robust structural priors, protecting the initial robust configurations from being degraded by later-stage fitting. Asymmetric Weight Reversion~(AWR) reverts the shallow subnetwork to an earlier robust state and stabilizes it during subsequent training, recovering robustness-relevant properties lost under continued training while leaving deeper layers free to adapt. We show that EPS and AWR improve robustness over competitive baselines across multiple corruption benchmarks, generalize across architectures, and are effective in downstream transfer, dynamic adaptation, and real-world physical scenarios. Beyond classification, they also yield gains on object detection, semantic segmentation, and continual test-time adaptation. Finally, ablations and analyses associate the gains with smoother loss geometry under input perturbations and more stable representations under distribution shift, together with improved shape bias and frequency-domain behavior. Our contributions are summarized as follows.
\begin{enumerate}
    \item We identify a robustness fading phenomenon: shallow layers spontaneously develop robust configurations and flatter loss landscapes in early training, but these properties are not preserved during standard convergence.
    \item We propose Early-Phase Stabilization and Asymmetric Weight Reversion to stabilize these early-emergent robust priors without architectural modifications or significant computational overhead.
    \item We demonstrate that our method outperforms competitive baselines, is applicable to diverse network architectures, and proves effective in downstream transfer, dynamic adaptation, and real-world physical scenarios.
\end{enumerate}


\section{Related Work}

\subsection{Robustness to Natural Corruptions} To enhance the robustness of deep neural networks against natural corruptions, various strategies have been explored, ranging from data augmentation (e.g., AugMix~\cite{hendrycksaugmix}, PRIME~\cite{modas2022prime}) and optimization (e.g., SAM~\cite{foret2020sharpness} and DAMP~\cite{trinh2024improving}) to ensembling (e.g., RoHL's multi-expert fusion~\cite{saikia2021improving}). Another line of research delves into layer-wise robustness by emphasizing the heterogeneous contributions of different network components. This approach typically involves identifying or reinforcing specific robust subnetworks (e.g., EWS~\cite{guo2022improving}, DST~\cite{wudynamic}, AdaSAP~\cite{bairadaptive}) to maintain stability under distribution shifts. In particular, existing literature highlights the pivotal role of front-end layers in defending against perturbations. For instance, VOneNet~\cite{dapello2020simulating} and handcrafted Gabor filters~\cite{perez2020gabor} introduce biologically-inspired structures, while RoHL~\cite{saikia2021improving} employs layer-wise TV regularization to explicitly enhance shallow-layer representations. Unlike these methods requiring rigid modifications or high overhead, we propose a plug-and-play framework that captures early-emerging robust priors through the evolution of shallow subnetworks without auxiliary costs.

\subsection{Training Dynamics and Model Robustness} 
Research on training dynamics reveals that deep networks undergo a distinct stage-wise learning process: initial phases determine final connectivity~\cite{achille2018critical,chimoto2024critical} and exhibit a spectral bias toward low-frequency, structural features~\cite{rahaman2019spectral} within specific linear subspaces~\cite{frankleearly}. This suggests an early emergence of robust, global representations. However, this robustness often deteriorates during late-stage convergence, a phenomenon termed robust overfitting~\cite{rice2020overfitting,yu2022understanding}. This degradation is typically attributed to feature competition~\cite{ilyas2019adversarial,hermann2020origins}, where models increasingly exploit non-robust statistical patterns (e.g., high-frequency textures) to minimize empirical risk, resulting in a sharpened loss landscape~\cite{li2018visualizing,izmailov2018averaging}. While prior studies offer extensive static analyses of natural robustness~\cite{geirhos2018imagenet,gavrikov2024can,liu2025comprehensive}, the layer-wise dynamical evolution during training remains under-explored. We define "Robustness Fading" to characterize the attenuation of early robust priors within shallow sub-networks, identifying a critical yet overlooked dimension of robustness fluctuations throughout the learning trajectory.


\begin{wrapfigure}{r}{0.48\textwidth}
  \centering
  \includegraphics[width=0.28\textwidth]{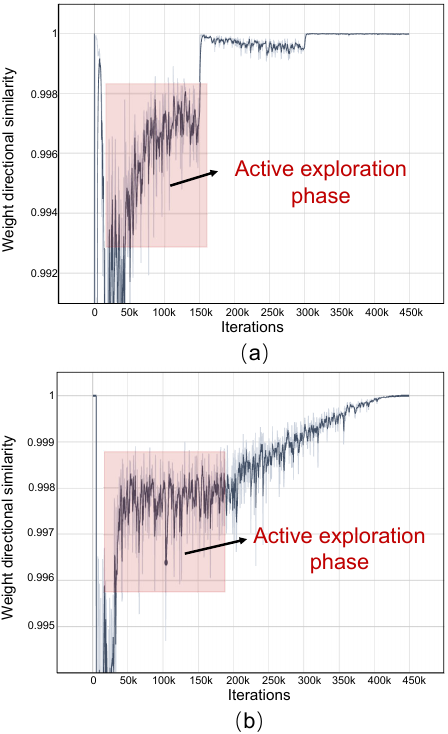}
  \caption{Weight directional stability evolution. (a) and (b) employ different LR schedulers. The shaded area highlights the initial unstable exploration phase.}
  \label{fig:wds_trajectory}
\end{wrapfigure}

\section{Early-emergent Robustness in Training Dynamics}

\subsection{Motivation}

As demonstrated in \cref{fig:robustness_gap_analysis}, layer-wise linear probing reveals that shallow layers possess a substantial, yet under-utilized, capacity for corruption robustness. Motivated by this observation, we investigate the evolution of this potential through the lens of training dynamics. Taking ResNet-18~\cite{he2016deep} as a representative case, we characterize the directional stability of shallow-layer weight trajectories by computing the cosine similarity between weights at consecutive iterations. \cref{fig:wds_trajectory} shows that shallow layers exhibit a prolonged active exploration phase early in training, with large step-to-step changes in weight direction rather than a quick stabilization. This indicates that early optimization drives substantial re-organization in shallow layers. We therefore hypothesize that robustness-relevant shallow patterns can emerge in this exploratory stage, but standard training fails to preserve them as optimization converges on clean data.

\subsection{Implicit Properties of Early-Phase Features}
\label{subsec: implicit_properties}
\begin{figure}[tb] 
  \centering
  \includegraphics[width=0.9\linewidth]{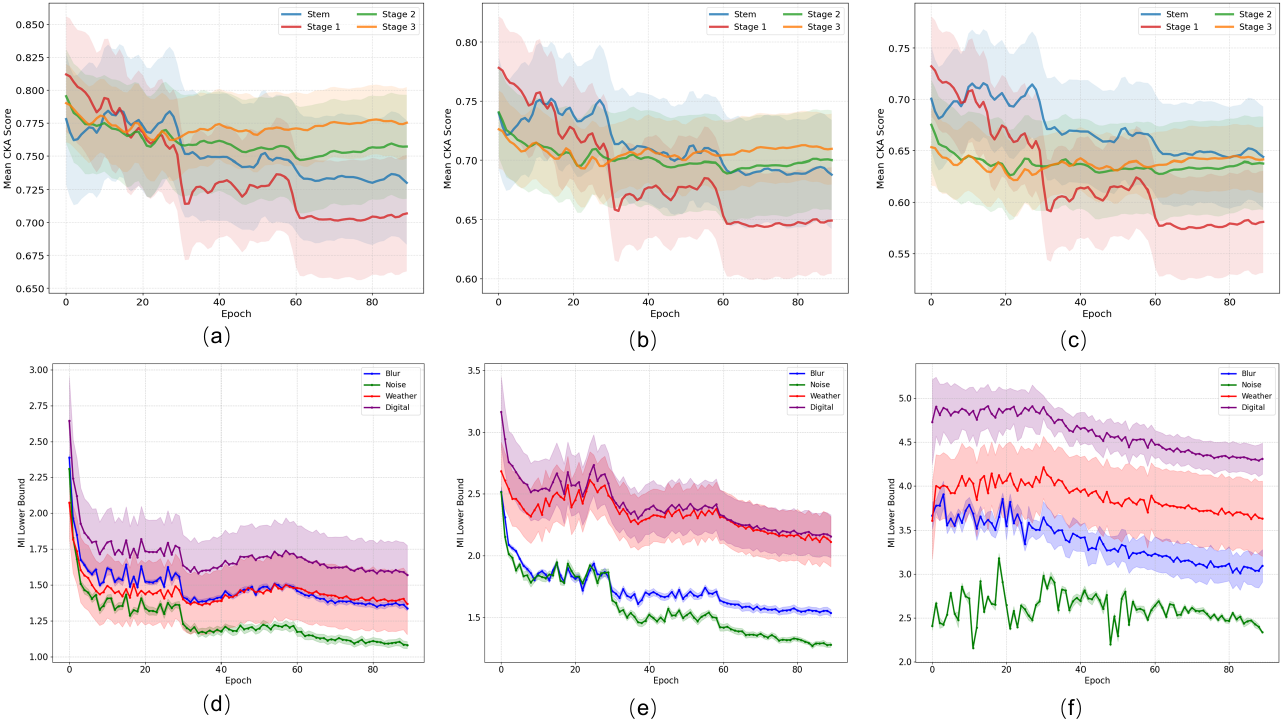} 
  \caption{Evolution of representation-level robustness across network depths and training phases. (a-c) CKA similarity between clean and corrupted features at severity levels 3, 4, and 5; each curve tracks a subnetwork's (Stem to Stage-3) stability during training. (d-f) InfoNCE-based MI estimates for Stem, Stage-1, and Stage-4 across four corruption types. Results reveal that shallow layers achieve peak structural similarity and information preservation in early training stages.}
  \label{fig:implicit properties}
\end{figure}
Following the observed parameter exploration phase, we further investigate robustness at the feature level using Centered Kernel Alignment (CKA)~\cite{kornblith2019similarity} and InfoNCE-based Mutual Information (MI)~\cite{oord2018representation}. By computing these metrics between features derived from clean and perturbed inputs, we quantify structural similarity and information retention across different subnetworks. As illustrated in \cref{fig:implicit properties}a-c, shallow subnetworks including the Stem and Stage-1 exhibit a progressive decline in CKA similarity. This downward trend is further intensified by higher corruption severities, whereas deeper stages maintain oscillating stability. This divergence is mirrored in MI dynamics (\cref{fig:implicit properties}d-f), where early-layer information preservation systematically decays while Stage-4 remains stagnant. These results suggest that shallow layers capture superior robust priors during the early training phase. However, such advantages are gradually washed away by subsequent optimization, highlighting a fundamental misalignment between standard training objectives and early-stage robust representations.
\subsection{Shaping the Flat Loss Landscape}
\begin{figure}[tb] 
  \centering
  \includegraphics[width=0.9\linewidth]{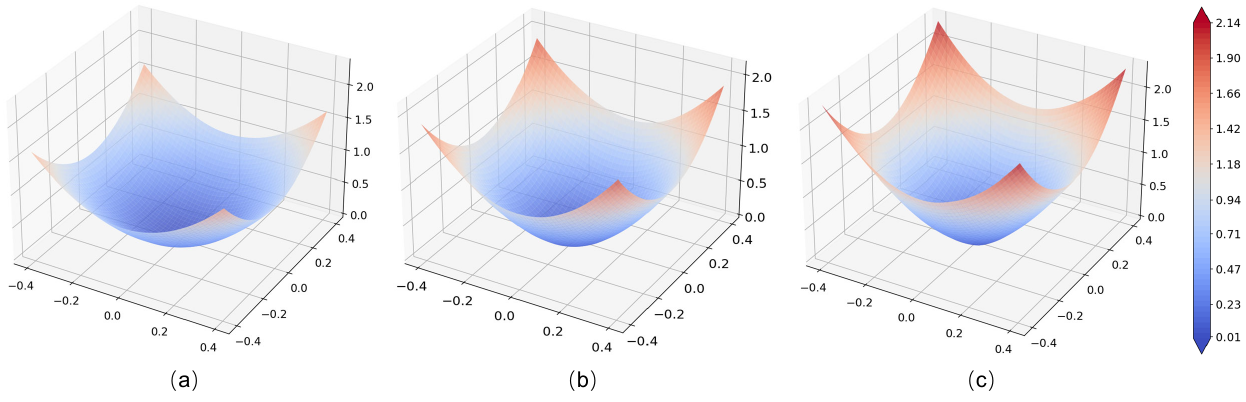} 
  \caption{Loss surfaces under perturbations, where shallow subnetworks are frozen at different training phases (early, intermediate, and late). Results suggest that early-phase shallow features possess an intrinsic property that guides the model toward flatter and more robust minima.}
  \label{fig:loss_landscape}
\end{figure}
To further explore shallow feature evolution, we shift from representation space to the geometric topology of the optimization landscape. We conduct causal freezing experiments by fixing shallow weights at different training epochs and re-optimizing the remaining layers, thereby isolating the influence of early-phase configurations on the model convergence. As demonstrated in \cref{fig:loss_landscape}, the loss curvature under perturbation exhibits distinct evolutionary paths depending on the timing of weight extraction. Specifically, models anchored by early-phase shallow weights yield a significantly smoother landscape with lower local sharpness, a property typically associated with superior robustness. In contrast, selecting weights from later stages induces a markedly steeper topography, a transition that persists despite subsequent retraining. This geometric shift reinforces our representation-level findings, indicating that robustness emerges early as a flat landscape but is systematically eroded during standard training. Such progressive sharpening highlights an inherent limitation of current optimization paradigms in preserving these transient yet foundational robust priors.
\section{Methodology: Interventions on Training Dynamics}
To bridge our empirical observations with a practical solution, we introduce a framework that treats robustness enhancement as a trajectory intervention problem. This section formalizes a unified perspective on how targeted constraints on shallow subnetworks can effectively prevent the systematic degradation of early-emergent robust priors.
\subsection{Preliminaries: A Unified View of Training Dynamics}
We begin by establishing a consistent notation for characterizing training trajectories. Consider a deep neural network parameterized by $\theta$. In the context of robustness against natural corruptions, we decompose the model parameters into two functional components:
\begin{equation}
\theta = \{\theta_s, \theta_d\}
\label{eq:param_decomposition}
\end{equation}
where $\theta_s$ denotes the shallow subnetwork responsible for low-level geometric and textural features, and $\theta_d$ represents the deep subnetwork dedicated to high-level semantic abstraction.During a standard training period $T$, parameter evolution follows Empirical Risk Minimization~(ERM), generating a trajectory $\{\theta^{(t)}\}_{t=1}^T$ that minimizes the loss $\mathcal{L}$. As established in our preceding analysis, the observed progressive sharpening of the loss landscape arises from the attenuation of robust priors in $\theta_s$ as $t \to T$. We thus formulate the intervention as an operator $\mathcal{I}$ that modulates the shallow updates starting at a specific time $\tau < T$:
\begin{equation}
\theta_s^{(t+1)} \leftarrow \mathcal{I}(\theta_s^{(t)}, \nabla_{\theta_s} \mathcal{L}, \dots), \quad \forall t \ge \tau
\label{eq:intervention_operator}
\end{equation}
This approach aims to mitigate the decay of robustness by explicitly constraining the temporal evolution of shallow parameters. Without altering the model architecture, $\mathcal{I}$ serves to preserve the early-emergent robust features discovered during the initial exploration phase.
\begin{algorithm}[t]
\caption{Interventions on Training Dynamics: EPS and AWR}
\label{alg:itd}
\footnotesize 
\begin{algorithmic}[1]
\REQUIRE Training set $\mathcal{D}$, Total epochs $T$, Initial $\theta^{(0)} = \{\theta_s^{(0)}, \theta_d^{(0)}\}$
\REQUIRE Intervention type $\mathcal{M} \in \{\text{EPS, AWR}\}$, Intervention epoch $\tau$, Snapshot epoch $t_{early}$
\REQUIRE Base learning rate $\eta$, Fine-tuning learning rate $\eta_{fine}$
\STATE \textbf{Initialize:} $S \leftarrow \emptyset$, $\eta_s \leftarrow \eta$, $\eta_d \leftarrow \eta$ 
\FOR{$t = 0$ \TO $T-1$}
    \STATE Sample mini-batch $(x, y) \sim \mathcal{D}$
    \STATE Compute loss $\mathcal{L} = \text{CrossEntropy}(\mathcal{F}(x; \theta^{(t)}), y)$
    
    \IF{$\mathcal{M} = \text{EPS}$}
        \IF{$t < \tau$}
            \STATE $\theta_s^{(t+1)} \leftarrow \theta_s^{(t)} - \eta_s \cdot \nabla_{\theta_s} \mathcal{L}$
        \ELSE
            \STATE $\theta_s^{(t+1)} \leftarrow \theta_s^{(t)}$ 
        \ENDIF
    \ELSIF{$\mathcal{M} = \text{AWR}$}
        \IF{$t = t_{early}$}
            \STATE $S \leftarrow \theta_s^{(t)}$ 
        \ENDIF
        \IF{$t = \tau$}
            \STATE $\theta_s^{(t)} \leftarrow S$ 
            \STATE $\eta_s \leftarrow \eta_{fine}$ 
        \ENDIF
        \STATE $\theta_s^{(t+1)} \leftarrow \theta_s^{(t)} - \eta_s \cdot \nabla_{\theta_s} \mathcal{L}$
    \ENDIF
    
    \STATE $\theta_d^{(t+1)} \leftarrow \theta_d^{(t)} - \eta_d \cdot \nabla_{\theta_d} \mathcal{L}$ 
\ENDFOR
\STATE \textbf{Return} Final parameters $\theta^{(T)}$
\end{algorithmic}
\end{algorithm}
\subsection{Early-Phase Stabilization}
Guided by our analysis of parameter trajectories, we propose Early-Phase Stabilization~(EPS). The core of EPS is a trajectory truncation mechanism applied to $\theta_s$. Since empirical evidence suggests that $\theta_s$ spontaneously develops robust representations and a flat loss landscape during the initial stages, truncating its evolution prevents the systematic degradation of these early features as the model continues its task fitting phase. Formally, within a general iterative optimization framework where $\theta^{(t+1)} = \theta^{(t)} + \Delta\theta^{(t)}$, EPS intervenes in the dynamics of $\theta_s$ by nullifying its updates:
\begin{equation}
\forall t \ge \tau_{eps}, \quad \Delta\theta_s^{(t)} \leftarrow 0
\label{eq:eps}
\end{equation}
where $\tau_{eps}$ denotes the critical intervention epoch. Following $\tau_{eps}$, the shallow parameters are explicitly frozen at their state $\theta_s^{(\tau_{eps})}$, while the deep parameters $\theta_d$ continue to be optimized for the remaining iterations. This design preserves the early-emergent robust priors while maintaining sufficient capacity in $\theta_d$ for task adaptation.  In practice, $\tau_{eps}$ is strategically selected to align with the period where $\theta_s$ exhibits high structural similarity and information retention, as analyzed in \cref{subsec: implicit_properties}, ensuring the intervention coincides with the period where $\theta_s$ possesses the most favorable robust properties.

\subsection{Asymmetric Weight Reversion}
While EPS preserves robustness through early intervention, we introduce Asymmetric Weight Reversion~(AWR) as a complementary strategy for active trajectory recovery. AWR draws inspiration from the weight rewinding mechanism in the Lottery Ticket Hypothesis~(LTH)~\cite{franklelottery}, which demonstrates that historical parameter states can serve as superior foundations for subsequent optimization. Extending this insight, AWR treats the training trajectory as a reversible path, allowing the model to explicitly backtrack $\theta_s$ to a prior state where it possessed stronger robust properties. Specifically, at a designated epoch $t = \tau_{awr}$, AWR resets the shallow parameters to a historical state:
\begin{equation}
\theta_s^{(\tau_{awr})} \leftarrow \theta_s^{(t_{early})}, \quad \text{where } t_{early} < \tau_{awr}
\label{eq:awr}
\end{equation}
Following the reversion, we modulate the update intensity of $\theta_s$ by significantly reducing its learning rate $\eta_s$ to maintain the stability of the recovered features. Meanwhile, the deep subnetwork $\theta_d$ continues its optimization to realign with the reverted feature backbone. This asymmetric update ensures that the shallow layers remain within a robust region of the parameter space while $\theta_d$ completes the remaining training objectives. By navigating back to a flatter optimization path, AWR systematically enhances the model’s defense against corruptions without any structural modifications.
\section{Experiments and Analysis}
\subsection{Experimental Setup}
\begin{table}[!t]
    \centering
    \caption{Comparison with representative baselines on ImageNet-C. We report Top-1 Accuracy~($\uparrow$). $^*$\ denotes used improved training recipe.}
    \small
    \resizebox{0.5\textwidth}{!}{
    \begin{tabular}{cccc}
    \hline
    \bf Baseline  &\bf DST  & \bf $\text{AdaSAP}_P^*$ &\bf EWS \\
    39.2 & 38.7~(-0.5) & 43.3~(+4.1) & 40.6~(+1.4)   \\
    \hline
    \bf SAM &\bf DAMP & \bf DAT  &\bf VOneNet \\
    39.8~(+0.6) & 41.4~(+2.2) & 41.1~(+1.9) & 40.3~(+1.1) \\
    \hline
     \bf Gabor Layers &\bf EPS & \bf AWR &\bf $\text{AWR}^*$ \\
     37.5~(-1.7) & 42.5~(+3.3) & 43.1~(+3.9) &  46.5~(+7.3) \\
    \hline
    \end{tabular}}
    \label{table baselines}
\end{table}
The proposed framework is evaluated on diverse robustness benchmarks and downstream tasks; full protocols and hyperparameters are in the Supplementary Material.

\textbf{Benchmarks and Metrics.} We evaluate our framework across eight robustness benchmarks, spanning image classification to downstream perception. For classification, we assess corruption robustness using ImageNet-C~\cite{hendrycks2019benchmarking},ImageNet-$\bar{C}$~\cite{mintun2021interaction} and ImageNet-3DCC~\cite{kar20223d} and ImageNetV2-C~\cite{recht2019imagenet}. To verify the model transferability, we extend evaluations to object detection (COCO-C~\cite{michaelis2019benchmarking}) and semantic segmentation (ADE20K-C, Cityscapes-C~\cite{kamann2021benchmarking}). All corrupted datasets are reserved for inference-only testing. We report standard performance (Top-1 Acc, mAP, or mIoU) alongside the mean Corruption Error (mCE)~\cite{hendrycks2019benchmarking} or average corruption performance. 

\textbf{Implementation Details.} Our base recipe employs a randomly initialized ResNet-50 trained on ImageNet~\cite{deng2009imagenet} for 90 epochs. We use SGD (0.9 momentum) with an initial learning rate of 0.1, decayed by 10\% at epochs 30 and 60, and a batch size of 256. To isolate our framework's effects, we exclude advanced augmentations like AugMix~\cite{hendrycksaugmix} or Mixup~\cite{zhang2018mixup}. Downstream tasks are implemented via MMDetection~\cite{chen2019mmdetection} and MMSegmentation~\cite{mmseg2020}. Our strategy integrates two components: EPS, which constrains the optimization trajectory during initial $T_{eps}$ epochs, and AWR, which rewinds specific convolutional weights to rectify late-stage dynamical deviations. Sensitivity analyses for EPS/AWR are in \cref{sec: ablation_studies}.

\subsection{Image Classification under Corruption}
\textbf{Comparison with Representative Baselines.} Using a ResNet-50 backbone, we compare our method against two categories of robust representation techniques: (1) Optimization-based constraints (SAM~\cite{foret2020sharpness}, DAMP~\cite{trinh2024improving}, DAT~\cite{mao2022enhance}) that regularize the training path, and (2) Structural priors, including dynamic sparse evolution (EWS~\cite{guo2022improving}, DST~\cite{wudynamic}, AdaSAP~\cite{bairadaptive}) and bio-inspired front-ends (VOneNet~\cite{dapello2020simulating}, Gabor layers~\cite{perez2020gabor}). Results in \cref{table baselines} show that our methods significantly outperform these baselines. Specifically, EPS and AWR achieve 42.5\% and 43.1\% Top-1 accuracy, respectively, surpassing SAM by up to 3.3\%. Notably, AWR with strong augmentation reaches 46.5\%, yielding a +7.3\% gain over the baseline. These results demonstrate the superior effectiveness of our interventions in enhancing model robustness against ImageNet-C corruptions.

\begin{table}[!t]
    \centering
    \caption{We report Top-1 Accuracy ($\uparrow$) and mean Corruption Error (mCE, $\downarrow$) across various backbones on ImageNet-100 and its variants (C, $\bar{\text{C}}$, 3DCC, and V2-C). Avg. mCE summarizes the performance across these benchmarks, while $(+)$ and $(-)$ denote changes relative to each backbone's baseline after applying EPS or AWR.}
    \small
    \resizebox{0.9\textwidth}{!}{
    \begin{tabular}{ccccccccc}
    \hline
    \textbf{Architecture} & \bf   & \bf IN-100 &\bf IN-100-C & \bf IN-100-$\bar{C}$ &\bf IN-100-3DCC &\bf IN-100V2-C & Avg.mCE~($\downarrow$) \\
    \hline
     \multirow{3}{*}{MobileNetV2} & \ding{55}  & 84.6 & 95.8 & 94.4 & 88.8 & 97.6 &  94.2 \\
      &\cellcolor{gray!20} EPS  & \cellcolor{gray!20} 84.2  &\cellcolor{gray!20} 92.8 &\cellcolor{gray!20} 93.9 &\cellcolor{gray!20} 87.8  &\cellcolor{gray!20} 96.0 &\cellcolor{gray!20}92.6~(\textbf{-1.6})  \\
      &\cellcolor{gray!20} AWR  & \cellcolor{gray!20} 84.2  &\cellcolor{gray!20} 93.4 &\cellcolor{gray!20} 94.1 &\cellcolor{gray!20} 87.8  &\cellcolor{gray!20} 95.6 &\cellcolor{gray!20}92.7~(\textbf{-1.5}) \\
     \multirow{3}{*}{WideResNet-50} & \ding{55}  & 85.6 & 88.3 & 88.4 & 83.3 & 92.1 &  88.0 \\
      &\cellcolor{gray!20} EPS  & \cellcolor{gray!20} 84.9  &\cellcolor{gray!20} 72.6 &\cellcolor{gray!20} 83.1 &\cellcolor{gray!20} 72.5  &\cellcolor{gray!20} 80.5 &\cellcolor{gray!20}77.2~(\textbf{-10.8})  \\
      &\cellcolor{gray!20} AWR  & \cellcolor{gray!20} 85.3 &\cellcolor{gray!20} 71.8 &\cellcolor{gray!20} 82.3  &\cellcolor{gray!20} 71.9 &\cellcolor{gray!20} 80.1 &\cellcolor{gray!20} 76.5~(\textbf{-11.5}) \\
     \multirow{3}{*}{MobileViT-S} & \ding{55}  & 85.4 & 88.8 & 92.5 & 84.8 & 92.9 & 89.8  \\
      &\cellcolor{gray!20} EPS  & \cellcolor{gray!20} 85.8  &\cellcolor{gray!20} 87.8 &\cellcolor{gray!20} 90.5 &\cellcolor{gray!20} 84.8  &\cellcolor{gray!20} 93.6 &\cellcolor{gray!20}89.2~(\textbf{-0.6})  \\
      &\cellcolor{gray!20} AWR  & \cellcolor{gray!20}  85.2 &\cellcolor{gray!20} 87.0 &\cellcolor{gray!20} 91.7  &\cellcolor{gray!20} 84.5 &\cellcolor{gray!20} 90.7 &\cellcolor{gray!20} 88.5~(\textbf{-1.3}) \\
     \multirow{3}{*}{EfficientFormer-L1} & \ding{55}  & 91.6 & 73.7 & 66.3 & 65.2 & 80.4 & 71.4 \\
      &\cellcolor{gray!20} EPS  & \cellcolor{gray!20} 92.5  &\cellcolor{gray!20} 68.0 &\cellcolor{gray!20} 62.9  &\cellcolor{gray!20} 60.0 &\cellcolor{gray!20} 75.8  &\cellcolor{gray!20} 66.7~(\textbf{-4.7}) \\
      &\cellcolor{gray!20} AWR  & \cellcolor{gray!20} 92.0  &\cellcolor{gray!20} 72.8 &\cellcolor{gray!20} 64.3  &\cellcolor{gray!20} 64.1 &\cellcolor{gray!20} 79.2 &\cellcolor{gray!20} 70.1~(\textbf{-1.3}) \\
     \multirow{3}{*}{Mambaout-femto} & \ding{55}  & 94.1 & 60.0 & 54.7 & 57.7 & 68.2 & 60.2 \\
      &\cellcolor{gray!20} EPS  & \cellcolor{gray!20} 93.4  &\cellcolor{gray!20} 57.2 &\cellcolor{gray!20}  50.8 &\cellcolor{gray!20} 52.9  &\cellcolor{gray!20} 66.5 &\cellcolor{gray!20} 56.9~(\textbf{-3.3}) \\
      &\cellcolor{gray!20} AWR  & \cellcolor{gray!20} 93.1  &\cellcolor{gray!20} 58.5 &\cellcolor{gray!20} 52.8  &\cellcolor{gray!20} 55.0 &\cellcolor{gray!20} 67.3 &\cellcolor{gray!20} 58.4~(\textbf{-1.8}) \\
    \hline
    \end{tabular}}
    \label{table_arch}
\end{table}

\begin{table}[!t]
    \centering
    \caption{Integration with data augmentation and regularization. We report Top-1 Accuracy ($\uparrow$) and Avg. mCE ($\downarrow$) on ImageNet-1K and its variants. Our methods (EPS and AWR) are integrated with representative baselines to evaluate their additive gains.}
    \small
    \resizebox{0.7\textwidth}{!}{
    \begin{tabular}{ccccccccc}
    \hline
    \textbf{Main} & \bf   & \bf IN &\bf IN-C & \bf IN-$\bar{C}$ &\bf IN-3DCC &\bf INV2-C & Avg.mCE~($\downarrow$) \\
    \hline
     \multirow{3}{*}{AugMix} & \ding{55}  & 76.1 & 71.7 & 74.0 & 69.6 & 78.4 &  73.4 \\
      &\cellcolor{gray!20} EPS  & \cellcolor{gray!20} 76.2  &\cellcolor{gray!20} 70.8 &\cellcolor{gray!20} 73.6 &\cellcolor{gray!20} 69.1 &\cellcolor{gray!20} 78.2 &\cellcolor{gray!20}72.9~(\textbf{-0.5})  \\
      &\cellcolor{gray!20} AWR  & \cellcolor{gray!20} 74.1  &\cellcolor{gray!20} 68.9 &\cellcolor{gray!20} 75.1  &\cellcolor{gray!20} 69.1 &\cellcolor{gray!20} 76.4 &\cellcolor{gray!20} 72.4~(\textbf{-1.0}) \\
     \multirow{3}{*}{AutoAug} & \ding{55}  & 76.4 & 73.2 & 76.9 & 69.6 & 79.8 & 74.9  \\
      &\cellcolor{gray!20} EPS  & \cellcolor{gray!20} 74.3  &\cellcolor{gray!20} 70.6 &\cellcolor{gray!20} 78.6 &\cellcolor{gray!20} 69.5 &\cellcolor{gray!20} 77.7 &\cellcolor{gray!20}74.1~(\textbf{-0.8})  \\
      &\cellcolor{gray!20} AWR  & \cellcolor{gray!20} 74.3  &\cellcolor{gray!20} 69.7 &\cellcolor{gray!20} 76.6  &\cellcolor{gray!20} 68.7 &\cellcolor{gray!20} 76.7 &\cellcolor{gray!20} 73.0~(\textbf{-1.9}) \\
     \multirow{3}{*}{CutMix} & \ding{55}  & 76.9 & 76.9 & 76.4 & 72.5 & 82.5 & 77.1 \\
      &\cellcolor{gray!20} EPS  & \cellcolor{gray!20} 74.3  &\cellcolor{gray!20} 73.4 &\cellcolor{gray!20} 76.9 &\cellcolor{gray!20} 72.4 &\cellcolor{gray!20} 79.9 &\cellcolor{gray!20} 75.7~(\textbf{-1.4}) \\
      &\cellcolor{gray!20} AWR  & \cellcolor{gray!20} 74.4  &\cellcolor{gray!20} 74.1 &\cellcolor{gray!20} 76.5  &\cellcolor{gray!20} 73.1 &\cellcolor{gray!20} 80.4 &\cellcolor{gray!20} 76.0~(\textbf{-1.1}) \\
     \multirow{3}{*}{Label Smoothing} & \ding{55}  & 76.6 & 75.2 & 77.1 & 72.1 & 81.5 & 76.5 \\
      &\cellcolor{gray!20} EPS  & \cellcolor{gray!20} 74.5 &\cellcolor{gray!20} 72.7 &\cellcolor{gray!20} 77.7  &\cellcolor{gray!20} 71.8 &\cellcolor{gray!20} 79.3 &\cellcolor{gray!20} 75.4~(\textbf{-1.1}) \\
      &\cellcolor{gray!20} AWR  & \cellcolor{gray!20} 74.7  &\cellcolor{gray!20} 72.6 &\cellcolor{gray!20} 77.0  &\cellcolor{gray!20} 71.9 &\cellcolor{gray!20} 79.3 &\cellcolor{gray!20} 75.2~(\textbf{-1.3}) \\
     \multirow{3}{*}{Dropout} & \ding{55}  & 76.3 & 76.4 & 78.6 & 72.9 & 82.4 & 77.6   \\
      &\cellcolor{gray!20} EPS  & \cellcolor{gray!20} 73.9  &\cellcolor{gray!20} 73.0 &\cellcolor{gray!20} 77.7  &\cellcolor{gray!20} 72.4 &\cellcolor{gray!20} 79.8 &\cellcolor{gray!20} 75.7~(\textbf{-1.9}) \\
      &\cellcolor{gray!20} AWR  & \cellcolor{gray!20} 74.0  &\cellcolor{gray!20} 74.1 &\cellcolor{gray!20} 78.8  &\cellcolor{gray!20} 73.0 &\cellcolor{gray!20} 80.6 &\cellcolor{gray!20} 76.6~(\textbf{-1.0}) \\
    \hline
    \end{tabular}}
    \label{table_da_mr}
\end{table}

\begin{table}[bt]
  \centering
  \begin{minipage}[t]{0.42\textwidth}
    \caption{Object detection on COCO-C. We report mAP ($\uparrow$) comparing backbones with our framework ($\checkmark$) against the baseline ($\times$).}
    \label{tab:det}
    \scriptsize 
    \begin{tabular}{cccc}
    \hline
    Architecture & \bf   & \bf COCO & \bf COCO-C \\
    \hline
     \multirow{2}{*}{FasterRCNN} & \ding{55}  & 37.6 & 17.5  \\
      &\cellcolor{gray!20} \ding{52}  & \cellcolor{gray!20} 37.0   &\cellcolor{gray!20} 17.9~(+0.4)  \\
     \multirow{2}{*}{YOLOv5} & \ding{55}  & 39.8  &  19.9   \\
      &\cellcolor{gray!20} \ding{52}  & \cellcolor{gray!20} 37.9  &\cellcolor{gray!20} 24.3~(+4.4)    \\
    \hline
    \end{tabular}
  \end{minipage}
  \hspace{0.5mm}
  \begin{minipage}[t]{0.56\textwidth}
    \caption{ Semantic segmentation results on ADE20K-C and Cityscapes-C. We report mIoU ($\uparrow$) on clean (\textcolor{gray}{\bf S}) and corrupted (\textcolor{gray}{\bf R}) data.}
    \label{tab:seg}
    \scriptsize 
    \begin{tabular}{cccccc}
    \hline
    \multirow{2}{*}{Architecture} & \bf   & \multicolumn{2}{c}{\bf ADE20K-C} &\multicolumn{2}{c}{\bf Cityscapes-C} \\
     & \bf   & \textcolor{gray}{\bf S}  & \textcolor{gray}{\bf R} & \textcolor{gray}{\bf S} & \textcolor{gray}{\bf R} \\
    \hline
     \multirow{2}{*}{DeepLabV3+} & \ding{55}  & 42.1 & 21.3 & 78.6 & 36.2 \\
      &\cellcolor{gray!20} \ding{52}   & \cellcolor{gray!20} 41.0  &\cellcolor{gray!20} 22.0~(+0.7) &\cellcolor{gray!20} 78.3 &\cellcolor{gray!20} 39.9~(+3.7) \\
    \hline
     \multirow{2}{*}{GCNet} & \ding{55}  & 40.3  & 19.9 & 76.8 & 33.1 \\
      &\cellcolor{gray!20} \ding{52}   & \cellcolor{gray!20} 39.0   &\cellcolor{gray!20} 20.7~(+0.8)  &\cellcolor{gray!20} 76.8 &\cellcolor{gray!20} 36.2~(+3.1) \\
    \hline
    \end{tabular}
  \end{minipage}
\end{table}

\begin{table}[bt]
  \centering
  \begin{minipage}[t]{0.4\textwidth}
    \caption{Online TTA performance on ImageNet-C (severity 5). We report Top-1 Accuracy ($\uparrow$).}
    \label{tab:tta}
    \scriptsize 
    \begin{tabular}{cccc}
    \hline
    & \bf Baseline  & \bf EPS & \bf AWR\\
    \hline
    ROTTA & 32.6 & 34.4 & 33.5\\
    AdaCont & 34.9 & 36.7 & 37.5  \\
    Tent & 37.3 &  37.4 & 37.9\\
    SAR & 37.8 & 37.9 & 38.3\\
    RMT & 42.2 & 43.8 & 43.7\\
    \hline
    \end{tabular}
  \end{minipage}
  \hspace{0.5mm}
  \begin{minipage}[t]{0.58\textwidth}
    \caption{Robustness evaluation on ACDC. We report mIoU ($\uparrow$).}
    \label{tab:acdc}
    \scriptsize 
    \begin{tabular}{ccccccc}
    \hline
    Architecture & \bf   & \bf Fog & \bf Night & \bf Rain & \bf Snow & \bf Avg. \\
    \hline
     \multirow{3}{*}{DeepLabV3+} & \ding{55}  & 62.1 & 14.6 & 48.0 & 44.8 & 42.4 \\
      &\cellcolor{gray!20} \ding{52}   & \cellcolor{gray!20} 62.3  &\cellcolor{gray!20} 15.5  &\cellcolor{gray!20} 47.2  &\cellcolor{gray!20} 44.9 &\cellcolor{gray!20} 42.5~(+0.1) \\
    \hline
     \multirow{3}{*}{GCNet} & \ding{55}  & 61.7 & 7.2 & 45.9 & 40.8 & 38.9\\
      &\cellcolor{gray!20} \ding{52}   & \cellcolor{gray!20} 61.8  &\cellcolor{gray!20} 8.8 &\cellcolor{gray!20} 45.1 &\cellcolor{gray!20} 42.8 &\cellcolor{gray!20} 39.6~(+0.7) \\
    \hline
    \end{tabular}
  \end{minipage}
\end{table}

\textbf{Versatility Across Diverse Architectures.} To demonstrate broad applicability, we evaluate various backbones. As shown in \cref{table_arch}, both EPS and AWR consistently reduce Average mCE. Specifically, EPS improves MobileNetV2~\cite{sandler2018mobilenetv2} (-1.6), EfficientFormer-L1~\cite{li2022efficientformer} (-4.7), and MambaOut-femto~\cite{yu2025mambaout} (-3.3), while AWR further suppresses mCE for WideResNet-50~\cite{zagoruyko2016wide} (-11.5) and MobileViT-S~\cite{mehta2021mobilevit} (-1.3). These gains confirm that our interventions provide an architecture-agnostic foundation for robust learning across CNN, ViT, and Mamba models.

\textbf{Synergy with Augmentation and Regularization.} We further evaluate the compatibility of EPS and AWR with representative data augmentation (AugMix~\cite{hendrycksaugmix}, AutoAug~\cite{cubuk2018autoaugment}) and regularization (Label Smoothing~\cite{muller2019does}, Dropout~\cite{srivastava2014dropout}, CutMix~\cite{yun2019cutmix}) techniques. As shown in \cref{table_da_mr}, our methods consistently yield additive gains across all baselines. Notably, AWR and EPS reduce Avg. mCE by up to 1.9 when integrated with AutoAug and Dropout, respectively. These results demonstrate that our framework provides a complementary optimization perspective to existing stochastic transformations, serving as a robust plug-and-play component for modern training pipelines.

\subsection{Beyond Classification: Robustness Transfer and Adaptation}
\textbf{Transferability to Downstream Tasks.} We evaluate pre-trained backbones on dense prediction to verify prior transferability. As shown in \cref{tab:det} and \cref{tab:seg}, our method consistently boosts performance: gaining +0.4\% (FasterRCNN~\cite{ren2016faster}) and +4.4\% (YOLOv5~\cite{jocher2022ultralytics}) mAP on COCO-C, and +3.7\% (DeepLabV3+~\cite{chen2018encoder}) and +0.8\% (GCNet~\cite{cao2019gcnet}) mIoU on Cityscapes-C and ADE20K-C. These cross-task improvements confirm that our robust representations effectively benefit fine-grained spatial tasks under degradation.
 
\textbf{Stability in Online Test-time Adaptation.} We assess our method in continual online TTA scenarios under ImageNet-C (severity 5) to examine its ability to mitigate model collapse. As shown in \cref{tab:tta}, integrating EPS or AWR into existing protocols—including ROTTA~\cite{yuan2023robust}, AdaCont~\cite{chen2022contrastive}, Tent~\cite{wangtent}, SAR~\cite{niu2023towards}, and RMT~\cite{dobler2023robust}—consistently improves Top-1 accuracy. Notably, AWR significantly boosts AdaCont (+2.6\%), while EPS yields substantial gains for ROTTA (+1.8\%) and RMT (+1.6\%). These results demonstrate that our framework effectively stabilizes optimization and enhances the compatibility of various TTA methods under severe distribution shifts.

\textbf{Robustness under Real-world Corruptions.} We further evaluate our framework on the ACDC~\cite{sakaridis2021acdc} segmentation dataset to test resilience against physical degradations. As shown in \cref{tab:acdc}, our framework ($\checkmark$) outperforms the baseline ($\times$) in average mIoU for both DeepLabV3+ (+0.1\%) and GCNet (+0.7\%). Despite a performance trade-off in the Rain scenario, substantial gains in Night and Snow demonstrate our method's efficacy in extreme visibility conditions.

\subsection{Ablation Studies}
\label{sec: ablation_studies}
\begin{figure}[tb] 
  \centering
  \includegraphics[width=0.98\linewidth]{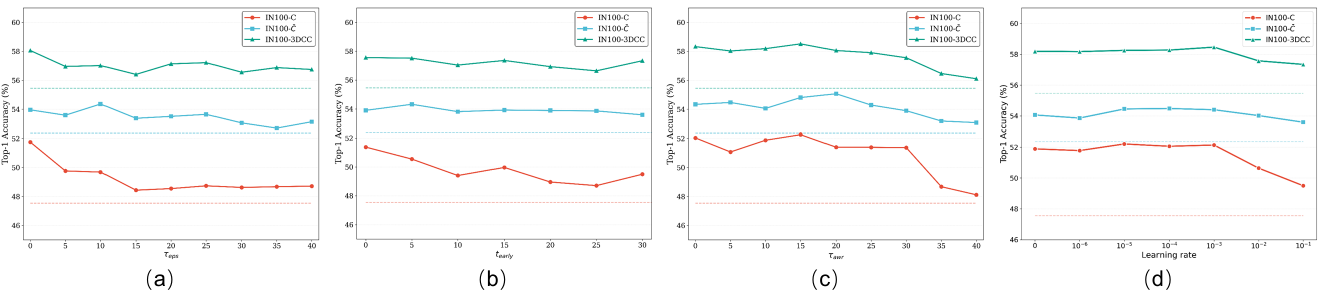} 
  \caption{Sensitivity analysis of (a) EPS intervention timing $\tau_{eps}$, (b) AWR hyper-parameters $t_{early}$, (c) $\tau_{awr}$, and (d) $\eta_s$.}
  \label{fig:ablation}
\end{figure}
\begin{figure}[tb] 
  \centering
  \includegraphics[width=0.98\linewidth]{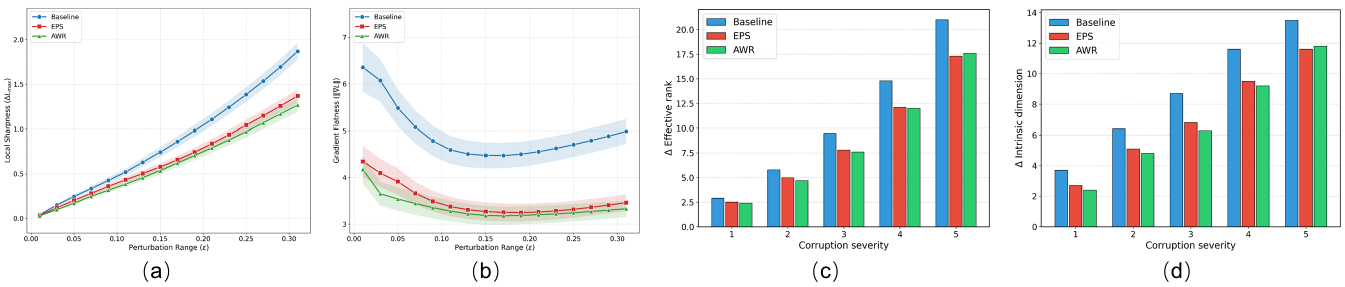} 
  \caption{Analysis of optimization sensitivity and representation stability. (a-b) Evolution of Local Sharpness ($\Delta L_{max}$) and Gradient Flatness ($\|\nabla L\|$) for Baseline, EPS, and AWR during training. (c-d) Representation geometry stability across five corruption severity levels, measured by the deviation in Effective Rank (ER) and Intrinsic Dimension (ID) between clean and corrupted samples.}
  \label{fig:analysis}
\end{figure}
\begin{figure}[tb] 
  \centering
  \includegraphics[width=0.9\linewidth]{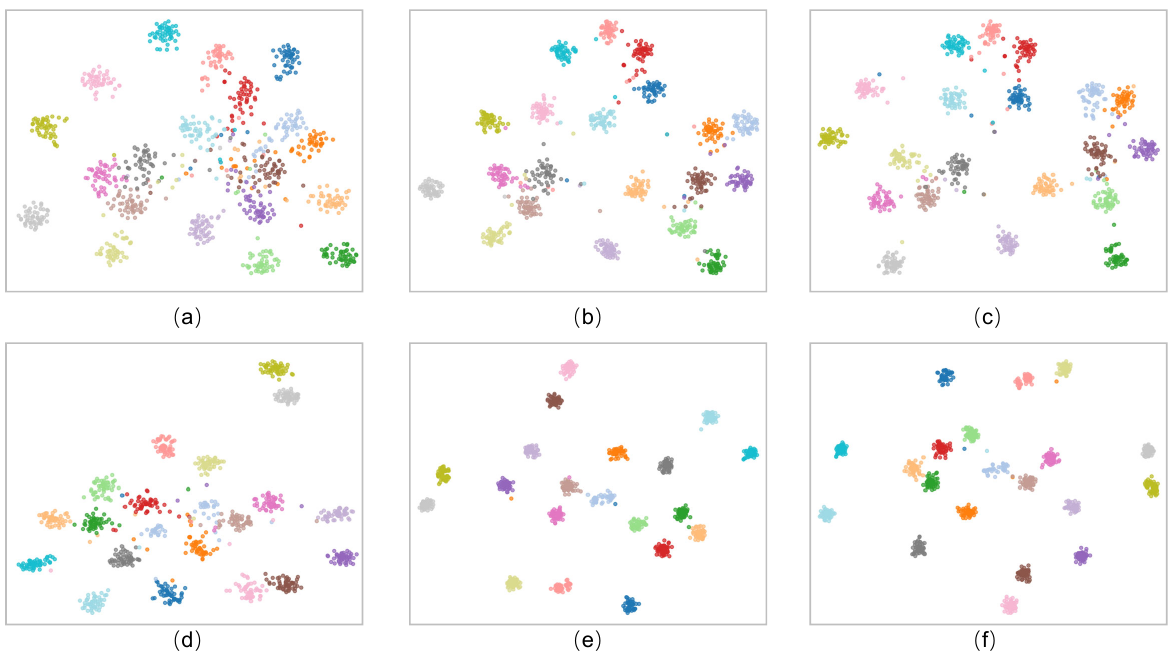} 
  \caption{t-SNE visualizations of Baseline, EPS, and AWR (from left to right) under (a-c) Gaussian noise and (d-f) Pixelate.}
  \label{fig:t_sne}
\end{figure}
\begin{figure}[tb] 
  \centering
  \includegraphics[width=0.6\linewidth]{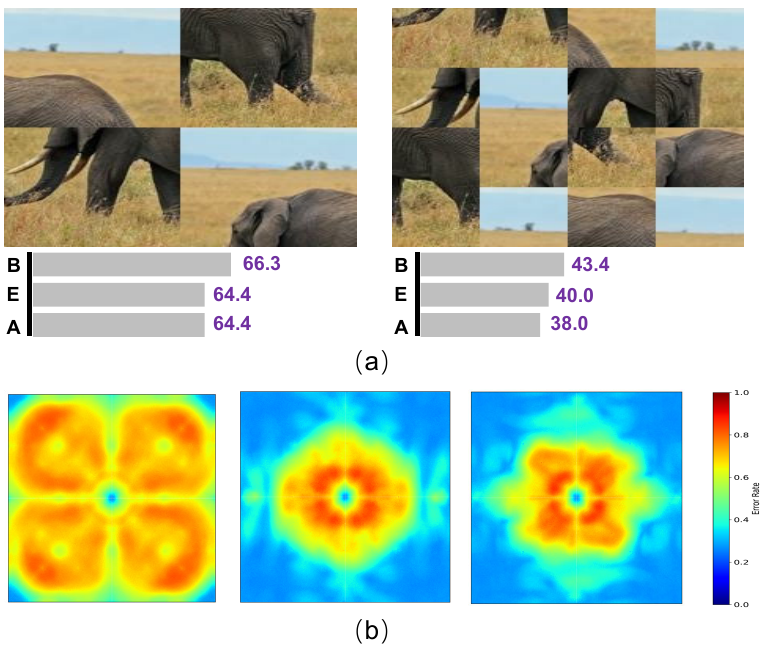} 
  \caption{Robustness analysis under patch shuffling and frequency domain perturbations. (a) Top-1 Accuracy under patch shuffling, grouped by patch sizes $2\times2$ and $4\times4$ (from left to right). (b) Frequency-domain error distribution for Baseline, EPS, and AWR (from left to right).}
  \label{fig:bias}
\end{figure}
\textbf{Early-Phase Stabilization Ablation.} We investigate the sensitivity of EPS to the intervention timing $\tau_{eps}$. \cref{fig:ablation}a illustrates the performance of ResNet-18 across IN100-C, IN100-$\bar{C}$, and IN100-3DCC as $\tau_{eps}$ varies from 0 to 40 epochs. As shown in \cref{fig:ablation}a, robustness exhibits a marginal downward trend as $\tau_{eps}$ increases. Nevertheless, EPS consistently maintains a performance lead over the baseline model across all tested intervals. This sustained superiority confirms that the efficacy of EPS is robust to the choice of intervention timing, demonstrating its practical stability during the early training phase.

\textbf{Asymmetric Weight Reversion Ablation.} We evaluate the synergy between $t_{early}$, $\tau_{awr}$, and $\eta_s$. As shown in \cref{fig:ablation}b, the performance of IN100-C, IN100-$\bar{C}$, and IN100-3DCC remains consistently superior to the dashed baselines across $t_{early} \in [0, 30]$, showing our method's insensitivity to the reversion anchor. For the trigger epoch \cref{fig:ablation}c, the accuracy peaks at $\tau_{awr} = 20$ and drops beyond 35, highlighting the need for balanced feature transition. Regarding \cref{fig:ablation}d, the model exhibits remarkable robustness within $[10^{-6}, 10^{-3}]$, with a sharp decline only at $10^{-1}$ due to optimization instability. Overall, these results confirm the practical reliability of the AWR mechanism across all benchmarks.
\subsection{Probing the Interventions}
\textbf{Optimization Geometry Sensitivity.}
To investigate the mechanisms of robustness enhancement, we analyze the loss landscape topology by perturbing inputs along random orthogonal bases across scales $\epsilon$, quantifying Local Sharpness ($\Delta L_{max}$)~\cite{santurkar2018does} and Gradient Flatness ($\|\nabla L\|$)~\cite{madry2018towards}. As shown in \cref{fig:analysis}, while the Baseline exhibits a rugged and sharp landscape, both EPS and AWR significantly stabilize the geometry. In \cref{fig:analysis}a, these interventions suppress the growth of $\Delta L_{max}$, ensuring reliability in worst-case scenarios, while the consistently lower gradient norms observed in \cref{fig:analysis}b indicate a clear transition from sharp to flat minima. This shift toward a smoother functional landscape directly correlates with the observed performance gains under image degradation, confirming that our interventions facilitate convergence to more robust local optima by maintaining a flatter loss surface across increasing perturbation scales.


\textbf{Stability of Representation Geometry.}
To understand how our interventions shape feature geometry, we employ two key metrics: Effective Rank (ER) \cite{roy2007effective} to assess feature concentration, and Intrinsic Dimension (ID) \cite{li2018measuring} to evaluate structural complexity. Specifically, we calculate the absolute discrepancy ($\Delta$) between metrics computed on clean and corrupted data. As shown in \cref{fig:analysis}c-d, while the baseline exhibits significant variance under corruption, EPS and AWR consistently minimize $\Delta$ in both metrics. This stability confirms that our framework effectively preserves structural consistency, preventing representation drift under heavy perturbations. This is qualitatively reinforced by t-SNE~\cite{van2008visualizing} visualizations in \cref{fig:t_sne}, where our methods yield more compact and discriminative clusters. The alignment between these quantitative metrics and visualization results demonstrates that EPS and AWR foster a robust representation geometry that remains invariant to data degradation.

\textbf{Robustness via Shape and Frequency Biases.}
We analyze robustness via shape~\cite{geirhos2018imagenet} and frequency biases~\cite{yin2019fourier}. In \cref{fig:bias}a, we disrupt spatial logic via patch shuffling ($2\times2, 4\times4$) to evaluate the reliance on global structures. The performance gap between our models (E/A) and the Baseline (B) suggests a shift from local textures to global geometric representations. This is corroborated by the frequency-domain error maps in \cref{fig:bias}b, where our methods significantly suppress error rates in mid-to-high frequency bands. By filtering non-semantic noise while preserving structural cues, our framework achieves superior stability against natural corruptions.

\section{Discussion}
In this work, we use ``robust priors'' to denote early-stage shallow representations that are empirically associated with corruption robustness, as characterized by representation similarity, information preservation, freezing interventions, and loss-landscape analyses. This interpretation motivates EPS and AWR as lightweight trajectory interventions that stabilize or recover such early-emergent properties without introducing additional learnable parameters. The mechanism is most directly aligned with robustness degradation caused by low-level visual corruptions, while broader distribution shifts may also involve semantic, contextual, or sensor-domain changes. Thus, preserving early robust priors should be viewed as a practical route to improving corruption robustness, rather than a complete solution to all forms of distribution shift.

\section{Conclusion}
This paper identifies the robustness fading phenomenon, where shallow layers spontaneously develop robust representations and flat loss landscapes early in training but lose these properties upon convergence. We introduce Early-Phase Stabilization and Asymmetric Weight Reversion as two parameter-free strategies to stabilize these early-emergent robust priors. Extensive empirical evaluations and geometric analyses associate these gains with smoother loss landscapes and improved representation stability under distribution shifts. Our framework consistently improves corruption robustness across diverse architectures and vision tasks, including detection and segmentation without architectural overhead. By shifting the focus from static inductive biases to temporal training dynamics, we provide a minimalist yet effective path toward building inherently reliable models.



%
%
\bibliographystyle{splncs04}
\bibliography{manuscript}

@String(ECCV  = {Eur. Conf. Comput. Vis.})

@String(NeurIPS = {Adv. Neural Inform. Process. Syst.})

@String(ECCV  = {ECCV})

@String(NeurIPS = {NeurIPS})

@inproceedings{hendrycksaugmix,
  title={AugMix: A Simple Data Processing Method to Improve Robustness and Uncertainty},
  author={Hendrycks, Dan and Mu, Norman and Cubuk, Ekin Dogus and Zoph, Barret and Gilmer, Justin and Lakshminarayanan, Balaji},
  booktitle={International Conference on Learning Representations}
}

@inproceedings{modas2022prime,
  title={Prime: A few primitives can boost robustness to common corruptions},
  author={Modas, Apostolos and Rade, Rahul and Ortiz-Jim{\'e}nez, Guillermo and Moosavi-Dezfooli, Seyed-Mohsen and Frossard, Pascal},
  booktitle={European Conference on Computer Vision},
  pages={623--640},
  year={2022},
  organization={Springer}
}

@article{foret2020sharpness,
  title={Sharpness-aware minimization for efficiently improving generalization},
  author={Foret, Pierre and Kleiner, Ariel and Mobahi, Hossein and Neyshabur, Behnam},
  journal={arXiv preprint arXiv:2010.01412},
  year={2020}
}

@inproceedings{wightman2021resnet,
  title={ResNet strikes back: An improved training procedure in timm},
  author={Wightman, Ross and Touvron, Hugo and Jegou, Herve},
  booktitle={NeurIPS 2021 Workshop on ImageNet: Past, Present, and Future}
}

@inproceedings{bairadaptive,
  title={Adaptive Sharpness-Aware Pruning for Robust Sparse Networks},
  author={Bair, Anna and Yin, Hongxu and Shen, Maying and Molchanov, Pavlo and Alvarez, Jose M},
  booktitle={The Twelfth International Conference on Learning Representations},
  year={2024}
}

@article{trinh2024improving,
  title={Improving robustness to corruptions with multiplicative weight perturbations},
  author={Trinh, Quoc Trung and Heinonen, Markus and Acerbi, Luigi and Kaski, Samuel},
  journal={Advances in Neural Information Processing Systems},
  volume={37},
  pages={35492--35516},
  year={2024}
}

@article{mao2022enhance,
  title={Enhance the visual representation via discrete adversarial training},
  author={Mao, Xiaofeng and Chen, Yuefeng and Duan, Ranjie and Zhu, Yao and Qi, Gege and Li, Xiaodan and Zhang, Rong and Xue, Hui and others},
  journal={Advances in Neural Information Processing Systems},
  volume={35},
  pages={7520--7533},
  year={2022}
}

@article{hendrycks2019benchmarking,
  title={Benchmarking neural network robustness to common corruptions and perturbations},
  author={Hendrycks, Dan and Dietterich, Thomas},
  journal={arXiv preprint arXiv:1903.12261},
  year={2019}
}

@article{michaelis2019benchmarking,
  title={Benchmarking robustness in object detection: Autonomous driving when winter is coming},
  author={Michaelis, Claudio and Mitzkus, Benjamin and Geirhos, Robert and Rusak, Evgenia and Bringmann, Oliver and Ecker, Alexander S and Bethge, Matthias and Brendel, Wieland},
  journal={arXiv preprint arXiv:1907.07484},
  year={2019}
}

@article{kamann2021benchmarking,
  title={Benchmarking the robustness of semantic segmentation models with respect to common corruptions},
  author={Kamann, Christoph and Rother, Carsten},
  journal={International journal of computer vision},
  volume={129},
  number={2},
  pages={462--483},
  year={2021},
  publisher={Springer}
}

@inproceedings{zhang2018mixup,
  title={mixup: Beyond Empirical Risk Minimization},
  author={Zhang, Hongyi and Cisse, Moustapha and Dauphin, Yann N and Lopez-Paz, David},
  booktitle={International Conference on Learning Representations},
  year={2018}
}

@inproceedings{saikia2021improving,
  title={Improving robustness against common corruptions with frequency biased models},
  author={Saikia, Tonmoy and Schmid, Cordelia and Brox, Thomas},
  booktitle={Proceedings of the IEEE/CVF International Conference on Computer Vision},
  pages={10211--10220},
  year={2021}
}

@article{diffenderfer2021winning,
  title={A winning hand: Compressing deep networks can improve out-of-distribution robustness},
  author={Diffenderfer, James and Bartoldson, Brian and Chaganti, Shreya and Zhang, Jize and Kailkhura, Bhavya},
  journal={Advances in neural information processing systems},
  volume={34},
  pages={664--676},
  year={2021}
}

@article{dapello2020simulating,
  title={Simulating a primary visual cortex at the front of CNNs improves robustness to image perturbations},
  author={Dapello, Joel and Marques, Tiago and Schrimpf, Martin and Geiger, Franziska and Cox, David and DiCarlo, James J},
  journal={Advances in Neural Information Processing Systems},
  volume={33},
  pages={13073--13087},
  year={2020}
}

@article{yin2019fourier,
  title={A fourier perspective on model robustness in computer vision},
  author={Yin, Dong and Gontijo Lopes, Raphael and Shlens, Jon and Cubuk, Ekin Dogus and Gilmer, Justin},
  journal={Advances in Neural Information Processing Systems},
  volume={32},
  year={2019}
}

@article{vryniotis2021train,
  title={How to train state-of-the-art models using torchvision’s latest primitives},
  author={Vryniotis, Vasilis},
  journal={PyTorch},
  year={2021}
}

@inproceedings{guo2022improving,
  title={Improving robustness by enhancing weak subnets},
  author={Guo, Yong and Stutz, David and Schiele, Bernt},
  booktitle={European Conference on Computer Vision},
  pages={320--338},
  year={2022},
  organization={Springer}
}

@article{mintun2021interaction,
  title={On interaction between augmentations and corruptions in natural corruption robustness},
  author={Mintun, Eric and Kirillov, Alexander and Xie, Saining},
  journal={Advances in Neural Information Processing Systems},
  volume={34},
  pages={3571--3583},
  year={2021}
}

@inproceedings{kar20223d,
  title={3d common corruptions and data augmentation},
  author={Kar, O{\u{g}}uzhan Fatih and Yeo, Teresa and Atanov, Andrei and Zamir, Amir},
  booktitle={Proceedings of the IEEE/CVF Conference on Computer Vision and Pattern Recognition},
  pages={18963--18974},
  year={2022}
}

@inproceedings{recht2019imagenet,
  title={Do imagenet classifiers generalize to imagenet?},
  author={Recht, Benjamin and Roelofs, Rebecca and Schmidt, Ludwig and Shankar, Vaishaal},
  booktitle={International conference on machine learning},
  pages={5389--5400},
  year={2019},
  organization={PMLR}
}

@inproceedings{he2016deep,
  title={Deep residual learning for image recognition},
  author={He, Kaiming and Zhang, Xiangyu and Ren, Shaoqing and Sun, Jian},
  booktitle={Proceedings of the IEEE conference on computer vision and pattern recognition},
  pages={770--778},
  year={2016}
}

@article{cubuk2018autoaugment,
  title={Autoaugment: Learning augmentation policies from data},
  author={Cubuk, Ekin D and Zoph, Barret and Mane, Dandelion and Vasudevan, Vijay and Le, Quoc V},
  journal={arXiv preprint arXiv:1805.09501},
  year={2018}
}

@article{muller2019does,
  title={When does label smoothing help?},
  author={M{\"u}ller, Rafael and Kornblith, Simon and Hinton, Geoffrey E},
  journal={Advances in neural information processing systems},
  volume={32},
  year={2019}
}

@inproceedings{chen2018encoder,
  title={Encoder-decoder with atrous separable convolution for semantic image segmentation},
  author={Chen, Liang-Chieh and Zhu, Yukun and Papandreou, George and Schroff, Florian and Adam, Hartwig},
  booktitle={Proceedings of the European conference on computer vision (ECCV)},
  pages={801--818},
  year={2018}
}

@inproceedings{cao2019gcnet,
  title={Gcnet: Non-local networks meet squeeze-excitation networks and beyond},
  author={Cao, Yue and Xu, Jiarui and Lin, Stephen and Wei, Fangyun and Hu, Han},
  booktitle={Proceedings of the IEEE/CVF international conference on computer vision workshops},
  pages={0--0},
  year={2019}
}

@inproceedings{deng2009imagenet,
  title={Imagenet: A large-scale hierarchical image database},
  author={Deng, Jia and Dong, Wei and Socher, Richard and Li, Li-Jia and Li, Kai and Fei-Fei, Li},
  booktitle={2009 IEEE conference on computer vision and pattern recognition},
  pages={248--255},
  year={2009},
  organization={Ieee}
}

@inproceedings{kornblith2019similarity,
  title={Similarity of neural network representations revisited},
  author={Kornblith, Simon and Norouzi, Mohammad and Lee, Honglak and Hinton, Geoffrey},
  booktitle={International conference on machine learning},
  pages={3519--3529},
  year={2019},
  organization={PMLR}
}

@article{oord2018representation,
  title={Representation learning with contrastive predictive coding},
  author={Oord, Aaron van den and Li, Yazhe and Vinyals, Oriol},
  journal={arXiv preprint arXiv:1807.03748},
  year={2018}
}

@article{chen2019mmdetection,
  title={MMDetection: Open mmlab detection toolbox and benchmark},
  author={Chen, Kai and Wang, Jiaqi and Pang, Jiangmiao and Cao, Yuhang and Xiong, Yu and Li, Xiaoxiao and Sun, Shuyang and Feng, Wansen and Liu, Ziwei and Xu, Jiarui and others},
  journal={arXiv preprint arXiv:1906.07155},
  year={2019}
}

@misc{mmseg2020,
    title={{MMSegmentation}: OpenMMLab Semantic Segmentation Toolbox and Benchmark},
    author={MMSegmentation Contributors},
    howpublished = {\url{https://github.com/open-mmlab/mmsegmentation}},
    year={2020}
}

@inproceedings{achille2018critical,
  title={Critical learning periods in deep networks},
  author={Achille, Alessandro and Rovere, Matteo and Soatto, Stefano},
  booktitle={International conference on learning representations},
  year={2018}
}

@inproceedings{frankleearly,
  title={The Early Phase of Neural Network Training},
  author={Frankle, Jonathan and Schwab, David J and Morcos, Ari S},
  booktitle={International Conference on Learning Representations}
}

@inproceedings{rahaman2019spectral,
  title={On the spectral bias of neural networks},
  author={Rahaman, Nasim and Baratin, Aristide and Arpit, Devansh and Draxler, Felix and Lin, Min and Hamprecht, Fred and Bengio, Yoshua and Courville, Aaron},
  booktitle={International conference on machine learning},
  pages={5301--5310},
  year={2019},
  organization={PMLR}
}

@inproceedings{yu2022understanding,
  title={Understanding robust overfitting of adversarial training and beyond},
  author={Yu, Chaojian and Han, Bo and Shen, Li and Yu, Jun and Gong, Chen and Gong, Mingming and Liu, Tongliang},
  booktitle={International Conference on Machine Learning},
  pages={25595--25610},
  year={2022},
  organization={PMLR}
}

@inproceedings{franklelottery,
  title={The Lottery Ticket Hypothesis: Finding Sparse, Trainable Neural Networks},
  author={Frankle, Jonathan and Carbin, Michael},
  booktitle={International Conference on Learning Representations}
}

@article{ilyas2019adversarial,
  title={Adversarial examples are not bugs, they are features},
  author={Ilyas, Andrew and Santurkar, Shibani and Tsipras, Dimitris and Engstrom, Logan and Tran, Brandon and Madry, Aleksander},
  journal={Advances in neural information processing systems},
  volume={32},
  year={2019}
}

@article{li2018visualizing,
  title={Visualizing the loss landscape of neural nets},
  author={Li, Hao and Xu, Zheng and Taylor, Gavin and Studer, Christoph and Goldstein, Tom},
  journal={Advances in neural information processing systems},
  volume={31},
  year={2018}
}

@inproceedings{izmailov2018averaging,
  title={Averaging weights leads to wider optima and better generalization},
  author={Izmailov, P and Wilson, AG and Podoprikhin, D and Vetrov, D and Garipov, T},
  booktitle={34th Conference on Uncertainty in Artificial Intelligence 2018, UAI 2018},
  pages={876--885},
  year={2018}
}

@article{liu2025comprehensive,
  title={A comprehensive study on robustness of image classification models: Benchmarking and rethinking},
  author={Liu, Chang and Dong, Yinpeng and Xiang, Wenzhao and Yang, Xiao and Su, Hang and Zhu, Jun and Chen, Yuefeng and He, Yuan and Xue, Hui and Zheng, Shibao},
  journal={International Journal of Computer Vision},
  volume={133},
  number={2},
  pages={567--589},
  year={2025},
  publisher={Springer}
}

@inproceedings{gavrikov2024can,
  title={Can biases in imagenet models explain generalization?},
  author={Gavrikov, Paul and Keuper, Janis},
  booktitle={Proceedings of the IEEE/CVF conference on computer vision and pattern recognition},
  pages={22184--22194},
  year={2024}
}

@inproceedings{geirhos2018imagenet,
  title={ImageNet-trained CNNs are biased towards texture; increasing shape bias improves accuracy and robustness},
  author={Geirhos, Robert and Rubisch, Patricia and Michaelis, Claudio and Bethge, Matthias and Wichmann, Felix A and Brendel, Wieland},
  booktitle={International conference on learning representations},
  year={2018}
}

@article{hermann2020origins,
  title={The origins and prevalence of texture bias in convolutional neural networks},
  author={Hermann, Katherine and Chen, Ting and Kornblith, Simon},
  journal={Advances in neural information processing systems},
  volume={33},
  pages={19000--19015},
  year={2020}
}

@inproceedings{rice2020overfitting,
  title={Overfitting in adversarially robust deep learning},
  author={Rice, Leslie and Wong, Eric and Kolter, Zico},
  booktitle={International conference on machine learning},
  pages={8093--8104},
  year={2020},
  organization={PMLR}
}

@inproceedings{chimoto2024critical,
  title={Critical learning periods: Leveraging early training dynamics for efficient data pruning},
  author={Chimoto, Everlyn Asiko and Gala, Jay and Ahia, Orevaoghene and Kreutzer, Julia and Bassett, Bruce A and Hooker, Sara},
  booktitle={Findings of the Association for Computational Linguistics: ACL 2024},
  pages={9407--9426},
  year={2024}
}

@article{jocher2022ultralytics,
  title={ultralytics/yolov5: v7. 0-yolov5 sota realtime instance segmentation},
  author={Jocher, Glenn and Chaurasia, Ayush and Stoken, Alex and Borovec, Jirka and Kwon, Yonghye and Michael, Kalen and Fang, Jiacong and Yifu, Zeng and Wong, Colin and Montes, Diego and others},
  journal={Zenodo},
  year={2022}
}

@article{niu2023towards,
  title={Towards stable test-time adaptation in dynamic wild world},
  author={Niu, Shuaicheng and Wu, Jiaxiang and Zhang, Yifan and Wen, Zhiquan and Chen, Yaofo and Zhao, Peilin and Tan, Mingkui},
  journal={arXiv preprint arXiv:2302.12400},
  year={2023}
}

@article{ren2016faster,
  title={Faster R-CNN: Towards real-time object detection with region proposal networks},
  author={Ren, Shaoqing and He, Kaiming and Girshick, Ross and Sun, Jian},
  journal={IEEE transactions on pattern analysis and machine intelligence},
  volume={39},
  number={6},
  pages={1137--1149},
  year={2016},
  publisher={IEEE}
}

@inproceedings{yun2019cutmix,
  title={Cutmix: Regularization strategy to train strong classifiers with localizable features},
  author={Yun, Sangdoo and Han, Dongyoon and Oh, Seong Joon and Chun, Sanghyuk and Choe, Junsuk and Yoo, Youngjoon},
  booktitle={Proceedings of the IEEE/CVF international conference on computer vision},
  pages={6023--6032},
  year={2019}
}

@inproceedings{perez2020gabor,
  title={Gabor layers enhance network robustness},
  author={P{\'e}rez, Juan C and Alfarra, Motasem and Jeanneret, Guillaume and Bibi, Adel and Thabet, Ali and Ghanem, Bernard and Arbel{\'a}ez, Pablo},
  booktitle={European Conference on Computer Vision},
  pages={450--466},
  year={2020},
  organization={Springer}
}

@article{zagoruyko2016wide,
  title={Wide residual networks},
  author={Zagoruyko, Sergey and Komodakis, Nikos},
  journal={arXiv preprint arXiv:1605.07146},
  year={2016}
}

@inproceedings{sandler2018mobilenetv2,
  title={Mobilenetv2: Inverted residuals and linear bottlenecks},
  author={Sandler, Mark and Howard, Andrew and Zhu, Menglong and Zhmoginov, Andrey and Chen, Liang-Chieh},
  booktitle={Proceedings of the IEEE conference on computer vision and pattern recognition},
  pages={4510--4520},
  year={2018}
}

@article{srivastava2014dropout,
  title={Dropout: a simple way to prevent neural networks from overfitting},
  author={Srivastava, Nitish and Hinton, Geoffrey and Krizhevsky, Alex and Sutskever, Ilya and Salakhutdinov, Ruslan},
  journal={The journal of machine learning research},
  volume={15},
  number={1},
  pages={1929--1958},
  year={2014},
  publisher={JMLR. org}
}

@inproceedings{zhao2022ood,
  title={Ood-cv: A benchmark for robustness to out-of-distribution shifts of individual nuisances in natural images},
  author={Zhao, Bingchen and Yu, Shaozuo and Ma, Wufei and Yu, Mingxin and Mei, Shenxiao and Wang, Angtian and He, Ju and Yuille, Alan and Kortylewski, Adam},
  booktitle={European conference on computer vision},
  pages={163--180},
  year={2022},
  organization={Springer}
}

@article{mayilvahanan2024does,
  title={Does CLIP’s generalization performance mainly stem from high train-test similarity?},
  author={Mayilvahanan, Prasanna and Wiedemer, Thadd{\"a}us and Rusak, Evgenia and Bethge, Matthias and Brendel, Wieland},
  year={2024}
}

@inproceedings{yu2025mambaout,
  title={Mambaout: Do we really need mamba for vision?},
  author={Yu, Weihao and Wang, Xinchao},
  booktitle={Proceedings of the Computer Vision and Pattern Recognition Conference},
  pages={4484--4496},
  year={2025}
}

@inproceedings{sakaridis2021acdc,
  title={ACDC: The adverse conditions dataset with correspondences for semantic driving scene understanding},
  author={Sakaridis, Christos and Dai, Dengxin and Van Gool, Luc},
  booktitle={Proceedings of the IEEE/CVF international conference on computer vision},
  pages={10765--10775},
  year={2021}
}

@article{alain2016understanding,
  title={Understanding intermediate layers using linear classifier probes},
  author={Alain, Guillaume and Bengio, Yoshua},
  journal={arXiv preprint arXiv:1610.01644},
  year={2016}
}

@inproceedings{chen2022contrastive,
  title={Contrastive test-time adaptation},
  author={Chen, Dian and Wang, Dequan and Darrell, Trevor and Ebrahimi, Sayna},
  booktitle={Proceedings of the IEEE/CVF Conference on Computer Vision and Pattern Recognition},
  pages={295--305},
  year={2022}
}

@article{santurkar2018does,
  title={How does batch normalization help optimization?},
  author={Santurkar, Shibani and Tsipras, Dimitris and Ilyas, Andrew and Madry, Aleksander},
  journal={Advances in neural information processing systems},
  volume={31},
  year={2018}
}

@inproceedings{madry2018towards,
  title={Towards Deep Learning Models Resistant to Adversarial Attacks},
  author={Madry, Aleksander and Makelov, Aleksandar and Schmidt, Ludwig and Tsipras, Dimitris and Vladu, Adrian},
  booktitle={International Conference on Learning Representations},
  year={2018}
}

@article{yuan2019adversarial,
  title={Adversarial examples: Attacks and defenses for deep learning},
  author={Yuan, Xiaoyong and He, Pan and Zhu, Qile and Li, Xiaolin},
  journal={IEEE transactions on neural networks and learning systems},
  volume={30},
  number={9},
  pages={2805--2824},
  year={2019},
  publisher={IEEE}
}

@inproceedings{roy2007effective,
  title={The effective rank: A measure of effective dimensionality},
  author={Roy, Olivier and Vetterli, Martin},
  booktitle={2007 15th European signal processing conference},
  pages={606--610},
  year={2007},
  organization={IEEE}
}

@inproceedings{li2018measuring,
  title={Measuring the Intrinsic Dimension of Objective Landscapes},
  author={Li, Chunyuan and Farkhoor, Heerad and Liu, Rosanne and Yosinski, Jason},
  booktitle={International Conference on Learning Representations},
  year={2018}
}

@article{van2008visualizing,
  title={Visualizing data using t-SNE.},
  author={Van der Maaten, Laurens and Hinton, Geoffrey},
  journal={Journal of machine learning research},
  volume={9},
  number={11},
  year={2008}
}

@inproceedings{wangtent,
  title={Tent: Fully Test-Time Adaptation by Entropy Minimization},
  author={Wang, Dequan and Shelhamer, Evan and Liu, Shaoteng and Olshausen, Bruno and Darrell, Trevor},
  booktitle={International Conference on Learning Representations},
  year={2021}
}

@inproceedings{dobler2023robust,
  title={Robust mean teacher for continual and gradual test-time adaptation},
  author={D{\"o}bler, Mario and Marsden, Robert A and Yang, Bin},
  booktitle={Proceedings of the IEEE/CVF Conference on Computer Vision and Pattern Recognition},
  pages={7704--7714},
  year={2023}
}

@inproceedings{vaish2024fourier,
  title={Fourier-basis functions to bridge augmentation gap: Rethinking frequency augmentation in image classification},
  author={Vaish, Puru and Wang, Shunxin and Strisciuglio, Nicola},
  booktitle={Proceedings of the IEEE/CVF Conference on Computer Vision and Pattern Recognition},
  pages={17763--17772},
  year={2024}
}

@inproceedings{yuan2023robust,
  title={Robust test-time adaptation in dynamic scenarios},
  author={Yuan, Longhui and Xie, Binhui and Li, Shuang},
  booktitle={Proceedings of the IEEE/CVF Conference on Computer Vision and Pattern Recognition},
  pages={15922--15932},
  year={2023}
}

@article{qin2022understanding,
  title={Understanding and improving robustness of vision transformers through patch-based negative augmentation},
  author={Qin, Yao and Zhang, Chiyuan and Chen, Ting and Lakshminarayanan, Balaji and Beutel, Alex and Wang, Xuezhi},
  journal={Advances in Neural Information Processing Systems},
  volume={35},
  pages={16276--16289},
  year={2022}
}

@inproceedings{wudynamic,
  title={Dynamic Sparse Training versus Dense Training: The Unexpected Winner in Image Corruption Robustness},
  author={Wu, Boqian and Xiao, Qiao and Wang, Shunxin and Strisciuglio, Nicola and Pechenizkiy, Mykola and van Keulen, Maurice and Mocanu, Decebal Constantin and Mocanu, Elena},
  booktitle={The Thirteenth International Conference on Learning Representations},
  year={2025}
}

@article{li2022efficientformer,
  title={Efficientformer: Vision transformers at mobilenet speed},
  author={Li, Yanyu and Yuan, Geng and Wen, Yang and Hu, Ju and Evangelidis, Georgios and Tulyakov, Sergey and Wang, Yanzhi and Ren, Jian},
  journal={Advances in Neural Information Processing Systems},
  volume={35},
  pages={12934--12949},
  year={2022}
}

@article{mehta2021mobilevit,
  title={Mobilevit: light-weight, general-purpose, and mobile-friendly vision transformer},
  author={Mehta, Sachin and Rastegari, Mohammad},
  journal={arXiv preprint arXiv:2110.02178},
  year={2021}
}
\end{document}